\pdfoutput=1

\documentclass[11pt]{article}

\usepackage{acl}
\usepackage{times}
\usepackage{latexsym}
\usepackage[T1]{fontenc}
\usepackage[utf8]{inputenc}
\usepackage{microtype}

\usepackage{graphicx}
\usepackage{multirow}
\usepackage{adjustbox}
\usepackage{tabularx}
\usepackage{booktabs}
\usepackage{amsmath}
\usepackage{amssymb}
\usepackage{hyperref}
\usepackage{bbm}
\usepackage{caption}
\usepackage{subcaption}
\usepackage{mdframed}
\usepackage{tabularx}
\usepackage{lipsum}
\usepackage{algorithm}
\usepackage[noend]{algorithmic}
\DeclareMathOperator*{\argmax}{argmax}
\usepackage{color,soul}
\usepackage{pifont}
\usepackage{inconsolata}

\usepackage{tikz}
\usetikzlibrary{shapes,arrows}

\newcommand{\cross}[0]{\textcolor[HTML]{DA3F32}{\bf \ding{55}}} 
\newcommand{\tick}[0]{\textcolor[HTML]{398D64}{\bf \ding{51}}}

\title{Strings from the Library of Babel: Random Sampling as a\\ Strong Baseline for Prompt Optimisation}

\author{Yao Lu$^{\dagger}$ \quad Jiayi Wang$^{\dagger}$ \quad Raphael Tang$^{\ddagger}$ \quad Sebastian Riedel$^{\dagger}$ \quad Pontus Stenetorp$^{\dagger}$\\
    $^{\dagger}$University College London \quad
    $^{\ddagger}$Comcast AI Technologies\\
    \texttt{\{yao.lu,s.riedel,p.stenetorp\}@cs.ucl.ac.uk} \\
    \texttt{ucabj45@ucl.ac.uk} \quad \texttt{raphael\_tang@comcast.com}
    }

\begin{document}
\maketitle
\begin{abstract}
Recent prompt optimisation approaches use the generative nature of language models to produce prompts -- even rivalling the performance of human-curated prompts.
In this paper, we demonstrate that randomly sampling tokens from the model vocabulary as ``separators'' can be \textit{as effective} as language models for prompt-style text classification.
Our experiments show that random separators are competitive baselines, having less than a 1\% difference compared to previous self-optimisation methods and showing a 12\% average relative improvement over strong human baselines across nine text classification tasks and eight language models. 
We further analyse this phenomenon in detail using three different random generation strategies, establishing that the language space is rich with potentially good separators, with a greater than 40\% average chance that a randomly drawn separator performs better than human-curated separators.
These observations challenge the common assumption that an effective prompt should be human readable or task relevant and establishes a strong baseline for prompt optimisation research.\footnote{Our implementation is publicly available at \url{https://github.com/yaolu/random-prompt}}
\end{abstract}

\section{Introduction}
Pre-trained large language models~\cite[PLMs,][]{peters2018deep,devlin2019bert,liu2019roberta,Radford2019LanguageMA-gpt2,touvron2023llama,touvron2023llama2, jiang2023mistral} have demonstrated remarkable performance when conditioned with appropriate context~\cite{petroni-etal-2019-language,petroni2020context,jiang2020can, shin2020autoprompt,davison2019commonsense}.
For instance, when given a query along with the phrase ``Let's think step by step,'' such models are capable of solving reasoning tasks~\cite{kojima2022large-zs-cot, wei2022chain-cot}.
These special tokens, often called ``\textit{separators}'', are usually placed at the end of the input data or at the beginning of the output~(Table~\ref{tab:intro-example}).

Recent work found that including thinking-style separators such as "take a deep breath" can significantly enhance reasoning performance~\cite{yang2023large-opro}.
Similarly, another study discovered that simply using ``SOLUTION:'' is even more effective~\cite{fernando2023promptbreeder}.
These findings suggest that the language space for separators might still be under-explored, with many effective options yet to be identified.
A common framework employed in this line of research involves starting with thinking-style separators, using a language model to generate alternatives, and then selecting effective separators based on certain criteria~\cite{zhou2022large-ape,yang2023large-opro,fernando2023promptbreeder,guo2023connecting-evoprompt}.
This optimisation-style framework has shown promise in automating the exploration of the language space, which addresses the challenge of relying on human expertise to develop task-specific separators.

\begin{figure}[!t]
    \centering
    \includegraphics[width=.95\linewidth]{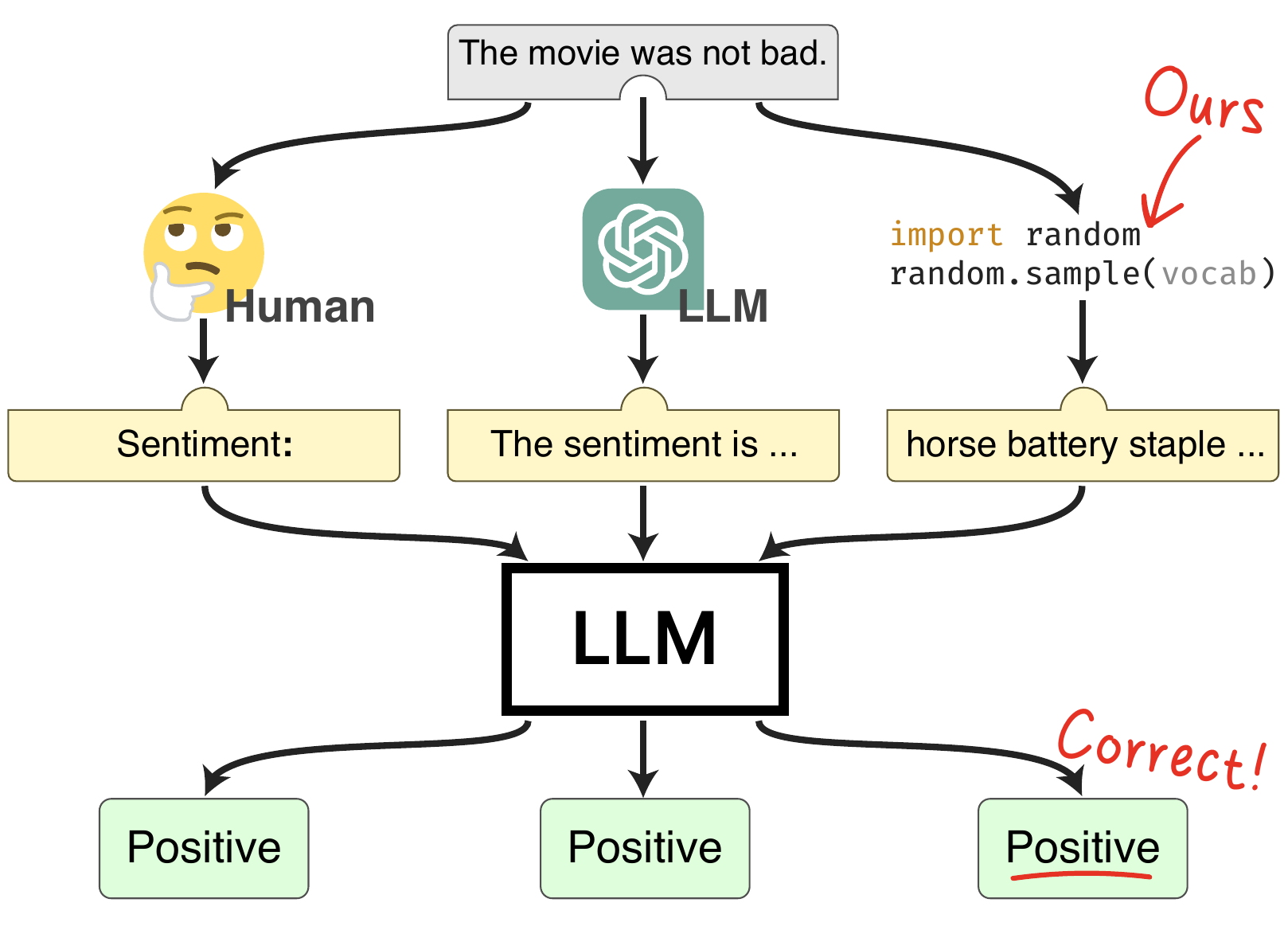}

    \caption{Illustration of our approach when searching for good separators for a sentiment classification task. Unlike relying on human knowledge or using external large language models to suggest alternatives, we find that randomly selected separators from the vocabulary can also yield good performance.}
    \label{fig:preface-curve}
\end{figure}

Most existing methods, particularly those applied to reasoning tasks, assume that effective separators should be closely related to the task or context~\cite{fernando2023promptbreeder, guo2023connecting-evoprompt,shi2022toward}.
However, perhaps counter-intuitively, we find that a performant separator does not necessarily have to be task-relevant or even coherent.
Sometimes, even tokens chosen at random from the vocabulary can improve performance as much as semantically meaningful phrases.
\begin{table}[!htb]
\setlength{\tabcolsep}{6pt}
\begin{center}
\centering
\resizebox{0.95\columnwidth}{!}{%
\begin{tabular}{ll}
\toprule
  & \textsc{Prompt Examples} \\
\midrule
\small \textsc{Human} & \small This is a good movie. \hl{Answer:} positive \\
\small \textsc{Random} & \small This is a good movie. \hl{!@\#?\&} positive \\
\bottomrule
\end{tabular}
}
\end{center}
\caption{
    Examples of prompts used in our evaluation, where the highlighted text are the separators.
}
\label{tab:intro-example}
\end{table}

As shown in Figure~\ref{fig:preface-curve} and Table~\ref{tab:intro-example}, we can achieve similar performance to that of human-optimised prompts by randomly selecting separators from the vocabulary. 
This suggests that random separators can serve as a competitive baseline, sometimes even matching the performance of previous methods~\cite{zhou2022large-ape, kojima2022large-zs-cot,  yang2023large-opro,guo2023connecting-evoprompt} for prompt-style text classification tasks.
Our exploration across seven different models further shows that this behaviour is universal across both pre-trained and instruction-tuned language models~(Table~\ref{tab:main_result}), and seems to be a fundamental characteristic of in-context learning~\cite{brown2020language-gpt3}.

We further analyse this phenomenon with three random separator generation strategies, revealing that there are many performant separators in the language space, suggesting that previous research underestimated the effectiveness of randomised prompts outside of reasoning tasks.
This observation breaks common assumptions such as that good separators need to be task relevant, coherent, and context dependent~\cite{shin2020autoprompt,shi2022toward}.
Experimental results show that using random separators attains a 12\% average relative improvement across nine classification tasks on eight language models, compared to human-curated separators.
To summarise, our contributions are as follows:
\begin{enumerate}
    \setlength\itemsep{-0em}
    \item{
        We show that random separators can be as effective as human-curated prompts for prompt-style text classification.
    }
    \item{
        We analyse three randomised separator generation strategies, which do not require an instruction-following language model, and show on-average a 12\% relative improvement over human baselines.
    }
    \item{
        We find that random separators are as competitive as previously proposed language model approaches to generate alternative prompts, suggesting that their effectiveness appear greater than what it would have appeared given this strong random baseline.
    }
\end{enumerate}

\section{Random Separator Optimisation}
\label{sec:random_separator}

\begin{figure*}
    \centering
    \includegraphics[width=0.94\linewidth]{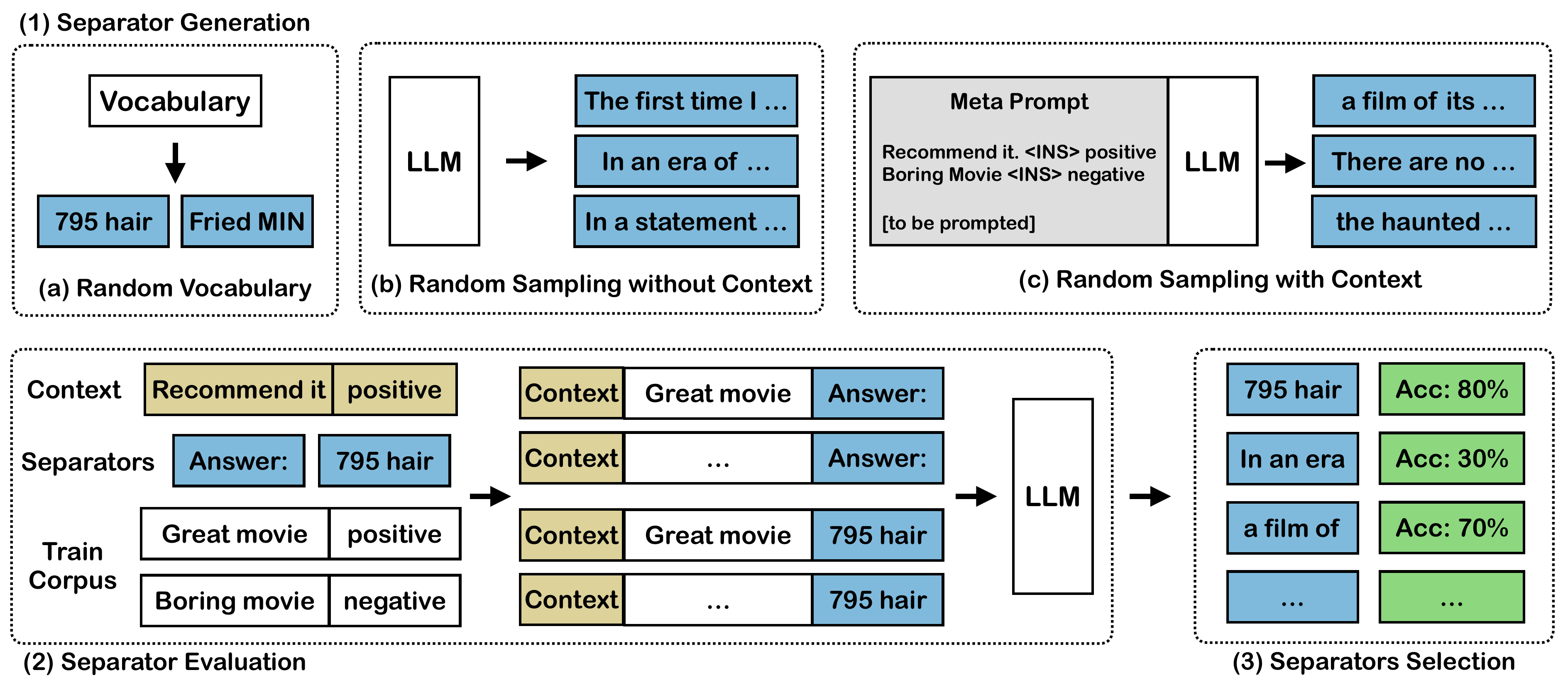}
    \caption{Our random separator optimisation procedure.}
    \label{fig:optimsation-flow}
\end{figure*}

We propose a random separator optimisation framework~(Figure~\ref{fig:optimsation-flow}) to find effective separators based on random sampling. The framework consists of three components: 1) random separator generation, 2) separator evaluation, and 3) separator selection. 

\subsection{Definition of Separator}
The term ``separators'' is inspired by the well-known BERT \texttt{[SEP]} token in order to differentiate from general wordings such as ``suffix'' or ``prefix''. We use this term to provide readers with better semantics and also to avoid ambiguity.

\subsection{Random Separator Generation}\label{sec:random-separator-generation}

In this work, we seek to answer the following questions which relate to common assumptions adopted in prompt optimisation research:
\begin{itemize}

\setlength\itemsep{-0.0em}
    \item{
        Should effective separators be task-relevant, or should they be closely tied to the existing context information?
    }
    \item{
        To what degree are language models needed for prompt optimisation?
    }
\end{itemize}
To answer these questions, we propose three strategies for generating random separators 
from language model free and context-free, to language model dependent and context relevant.
The core difference between these three random strategies are summarised in Table~\ref{tab:three-random-method-compare}, with a more detailed illustration provided in Table~\ref{tab:prompt_type}.

\begin{table}[htb]
\begin{center}
\centering
\resizebox{1.0\columnwidth}{!}{
\renewcommand{\arraystretch}{1.0} 

\begin{tabular}{lcc}
\toprule
 &  Language Model & Task Relevant\\
\midrule
Random Vocabulary & \cross & \cross \\ 
Random w/o Context & \tick & \cross \\
Random with Context & \tick & \tick \\
\bottomrule
\end{tabular}
}
\end{center}
\caption{Random generation methods} 
\label{tab:three-random-method-compare}
\end{table}

\paragraph{Sampling randomly across the vocabulary.} AutoPrompt~\cite{shin2020autoprompt} suggests that effective separators may appear to be random strings, showing the potential of identifying a good separator within the random space. To investigate this, we design an approach to generate random separators that are context-free, task-agnostic, and do not rely on any language model for their generation. Essentially, we select random tokens from the vocabulary until a predetermined string length limit is reached.

\paragraph{Sampling from a language model without context.} This method involves drawing samples from the language model's prior distribution, a process that is both context-free and task-agnostic. We employ this random generation method to evaluate whether separator coherence contributes to performance, especially when compared to random samples at the vocabulary level.

\paragraph{Sampling from a language model with context.} The creation of task-relevant separators may enhance performance. For instance, incorporating thinking-style phrases in a reasoning task has proven highly effective. In OPRO~\cite{yang2023large-opro}, the authors highlight that including samples from the training data in the meta-prompt leads to consistent improvements. Thus, to assess the impact of task relevance on separator generation, we integrate a few training samples from the training corpus into the meta-prompt to refine the semantic space of the separator.

\subsection{Separator Evaluation}

We demonstrate the evaluation process in Figure~\ref{fig:optimsation-flow}. In line with prior work~\cite{fernando2023promptbreeder, yang2023large-opro}, a small set of labelled data, denoted as the training corpus $ T = \{(x_i, y_i)\}, i=1,...,n $, is available. Here, $x_i$ and $y_i$ represent the sentence and label of the $i^\text{th}$ sample, respectively. We also define a transformation $\mathcal{T}$, which maps label $y_i$ to text.
In contrast to supervised learning settings that require a large volume of data for training, we only need a limited set of examples.\footnote{For all experiments, the training set size is $n=64$, unless explicitly mentioned otherwise.}
To evaluate a given separator $s$, we perform string concatenation ($\oplus$) of the each input sentence from the training corpus $T$ with the separator. 
As part of in-context learning settings, where demonstrations may be necessary for some tasks, we also take into account a context $c$, which has identical structure to the linearised sequence.
For each data point $(x_i, y_i)$ in $T$, we prompt the pretrained language model to generate the predicted label $\hat{y}_i = \argmax_{v \in V} P (v| c \oplus x_i \oplus s; \theta)$, where $\theta$ represents the parameters of the pretrained language model and $V$ denotes the vocabulary space.
For a classification task, we compute the classification accuracy as the corresponding separator score, denoted by $m$.
It is worth noting that the metric does not necessarily need to be differentiable, allowing for the direct optimisation of discrete metrics such as word overlap ratio, etc.

\subsection{Separator Selection}

Despite the random separator generation steps being independent, we retain the term ``iteration'' in our methodology. This allows for a consistent comparison with other methods that use an iteration-sensitive meta-prompt.
During the iteration process, we evaluate the generated prompt and keep track of the most effective separators at each step. The sampling continues until we reach a predefined sampling budget limit, denoted as $k$. By the end of the training process, we accumulate a set of separators\footnote{We generate up to 160 separators in all experiments.} and their corresponding scores, represented as $S = \{(s_i, m_i)\}, i = 1, ..., k$. 
The separator that yields the highest score in this set is then selected for evaluation on the test set.

\begin{table*}[!t]
\centering
\resizebox{2.0\columnwidth}{!}{%
\begin{tabular}{lll}
\toprule
 &
  Description &
  Prompt for Generating New Separators \\ \midrule
Random Baseline &
  Random string without optimisation steps&-
   \\
Human Baseline~\cite{lu-etal-2022-fantastically} &
Use ``{Answer:}'' as it is widely used &
  - \\
ZS-CoT~\cite{kojima2022large-zs-cot} &
  Think-style phrase &-
   \\ \hline
OPRO~\cite{yang2023large-opro} &
  \begin{tabular}[c]{@{}l@{}} The meta-prompt in OPRO consists of a\\natural language problem description and \\ instructions to generate new solutions based \\on previously found solutions. 
  \end{tabular} &
  \begin{tabular}[c]{@{}l@{}}{[}Instructions{]} I have some texts along with their ...\\ {[}Historical Solutions{]}  text: think stepwise   score: 55 ... \\ {[}Instructions{]} The following exemplars show ...\\ {[}Context{]} should not be missed \textless{}INS\textgreater positive ...\\ {[}Instructions{]} Write your new text that is different  ...\\ text: {[}to be prompted{]}\end{tabular} \\\hline
OPRO-ICL &

\begin{tabular}[c]{@{}l@{}} We remove all instructions from OPRO \\to create an in-context learning variant. 
  \end{tabular}
&
  \begin{tabular}[c]{@{}l@{}}{[}Historical Solutions{]}  text: think stepwise   score: 55 ... \\ {[}Context{]} should not be missed \textless{}INS\textgreater positive ...\\ text: {[}to be prompted{]}\end{tabular} \\  \hline
Random Vocabulary &

\begin{tabular}[c]{@{}l@{}} Randomly sample tokens across vocabulary \\to generate separators.
  \end{tabular}
 &
  \begin{tabular}[c]{@{}l@{}}{[}to be sampled{]}  {[}such as “!@\#\$\%\textasciicircum{}\&*”{]} \end{tabular} \\ \hline
Random w/o Context &
  Draw samples from LLM’s prior distribution. &
  {[}to be prompted{]} \\ \hline
Random with Context &
\begin{tabular}[c]{@{}l@{}}   Prompting language model with few examples \\as context. Similar to OPRO~\cite{yang2023large-opro},\\ we use three randomly sampled examples from\\ training data as context. 
  \end{tabular}
 &
  \begin{tabular}[c]{@{}l@{}}{[}Context{]}\\ should not be missed \textless{}INS\textgreater positive\\curiously depressing \textless{}INS\textgreater negative\\text: {[}to be prompted{]}
  \end{tabular}  \\ \hline

\end{tabular}%
}
\caption{Separator generation methods used for our main experiment. Text wrapped with square brackets are not included in prompt. 
}
\label{tab:prompt_type}
\end{table*}

\section{Experimental Setup}

\subsection{Datasets and Models}

In line with previous studies~\cite{gao2020making,zhao2021calibrate,lu-etal-2022-fantastically}, we use nine text classification datasets~(Table~\ref{tab:datasets}). For training we use 64 samples per dataset and for evaluation we use the sub-sampled test set from~\citet{lu-etal-2022-fantastically}.

\begin{table}[htb]
\begin{center}
\centering
\resizebox{1.0\columnwidth}{!}{
\renewcommand{\arraystretch}{1.0}

\begin{tabular}{lrrc}
\toprule
Dataset &  $\#$ of Classes & Avg. Len. & Balanced\\
\midrule
SST-2~\cite{socher2013sst2} & 2 & 12.4 & \tick\\
SST-5~\cite{socher2013sst2} & 5 & 23.1 & \cross \\
MR~\cite{pang2005mr} & 2 & 25.7 & \tick\\
CR~\cite{hu2004cr} & 2 & 22.1 & \tick\\
MPQA~\cite{wiebe2005mpqa} & 2 & 3.9 & \tick\\
Subj~\cite{pang2004subj} & 2 & 28.9 & \tick\\
TREC~\cite{voorhees2000trec} & 6 & 11.6 & \cross\\
AGNews~\cite{zhang2015agnewsdbpedia} & 4 &  53.8 & \tick\\
DBPedia~\cite{zhang2015agnewsdbpedia} & 14 & 65.5 & \tick\\
\bottomrule
\end{tabular}
}
\end{center}
\caption{Statistics of evaluation datasets, average length is calculated based on GPT-2 sentence-piece length.} 
\label{tab:datasets}
\end{table}
\begin{table}[htb]
\begin{center}
\centering
\resizebox{1.0\columnwidth}{!}{
\renewcommand{\arraystretch}{1.0} 

\begin{tabular}{lrc}
\toprule
\multirow{2}{*}{Model} & \multirow{2}{*}{\# of Parameters}    & \multicolumn{1}{c}{\multirow{2}{*}{\begin{tabular}[c]{@{}c@{}}Instruction \\ Tuned\end{tabular}}}   \\
&     & \multicolumn{1}{c}{}  \\
 \midrule

GPT2 Large~\cite{Radford2019LanguageMA-gpt2}                     & 0.8B       &     \cross            \\
GPT2 XL~\cite{Radford2019LanguageMA-gpt2}                        & 1.5B       &     \cross            \\
Mistral 7B ~\cite{jiang2023mistral}                        & 7B       &     \cross            \\
Mistral 7B Instruct~\cite{jiang2023mistral}                        & 7B       &     \tick            \\

Llama-Alpaca 7B~\cite{alpaca}                       & 6.7B       &     \tick         \\
Llama2 7B~\cite{touvron2023llama2}                       & 6.7B       &     \cross         \\
Llama2 7B Chat~\cite{touvron2023llama2}                   & 6.7B       &  \tick                \\
ChatGPT (GPT-3.5 Turbo, 0613 version)                    & --          &  \tick
   
\\
\bottomrule
\end{tabular}
}
\end{center}
\caption{Language models used in experiments.}
\label{tab:baseline-models}
\end{table}

In contrast to previous work, our random separator optimisation methods do not require an instruction-tuned language model. This allows us to test our methods using both standard pre-trained language models and instruction-tuned language models.
As detailed in Table~\ref{tab:baseline-models}, our experiment uses four pre-trained language models and four instruction-tuned language models, in total eight models with varying structure and training data. 

\subsection{Optimisation Settings}

\paragraph{Separator generation methods.} As detailed in Section~\ref{sec:random_separator}, we have proposed three different random separator generation approaches. For comparative analysis, we include the OPRO~\citep{yang2023large-opro} method in our study. We also adapt OPRO's meta-prompt by omitting the instructional text, creating an in-context learning variant~(OPRO-ICL). This allows fair comparison between random generation and other methods on non-instructionally tuned models.
We also compare our methods against two human-level prompt
optimisation methods, MI~\citep{zhang2023sentiment-mi} and NI~\citep{mishra2021cross-ni}, as well as two automatic prompt
optimisation methods, APE~\citep{zhou2022large-ape} and EvoPrompt~\citep{guo2023connecting-evoprompt}, under our experimental settings\footnote{Due to the difficulty of accurately reproducing these methods, we adapt our settings to fit the EvoPrompt setup instead.} described in \citet{guo2023connecting-evoprompt}. %
In total, we employ nine distinct separator generation methods as baselines in our main experiment~(Tables~\ref{tab:main_result} and \ref{tab:more-models}).

\paragraph{Baseline separators.} To better understand how much improvement separator optimisation can achieve, we use the human curated separator ``{Answer:}'' and random strings such as ``{Foo Bar}''\footnote{Here, ``{Foo Bar}'' represents a random string sampled from the vocabulary and not the literal usage of this exact string.} and zero-shot chain-of-thoughts~(ZS-CoT)~\cite{kojima2022large-zs-cot}, ``{Let's think step by step}'', as baseline separators. 

\paragraph{Initialisation.}
The choice of a starting point does not affect random methods. In cases where a meta-prompt requires a starting point, such as in OPRO, we use ``Answer:'' as the starting point.

\newcommand{\emphasize}[1]{\textcolor{blue}{#1}}

\begin{table*}[!t]
\begin{center}
\centering
\resizebox{2.0\columnwidth}{!}{
\begingroup
\renewcommand{\arraystretch}{0.80}
\begin{tabular}{lccccccccccc}
\toprule[1pt]
      \multirow{2}{*}{~} &
  \multirow{2}{*}{SST-2} &
  \multirow{2}{*}{SST-5} &
  \multirow{2}{*}{DBPedia} &
  \multirow{2}{*}{MR} &
  \multirow{2}{*}{CR} &
  \multirow{2}{*}{MPQA} &
  \multirow{2}{*}{Subj} &
  \multirow{2}{*}{TREC} &
  \multirow{2}{*}{AGNews} &
  \multirow{2}{*}{\begin{tabular}[c]{@{}c@{}}Avg. \\ (Rel. $\Delta$\%)\end{tabular}} \\
 &
   &
   &
   &
   &
   &
   &
   &
   &
   &
   \\
\midrule[1pt]

\small  Finetuning~(Full) & 95.0 & 58.7 &  99.3 & 90.8 & 89.4 &  87.8 & 97.0 & 97.4 & 94.7 & 90.0 \\

\midrule

\small \textsc{GPT2-Large 0.8B}   & & & & & & & & & &\\

~ \small \texttt{Answer:}  & $74.8$ &
$27.0$ &
$37.5$ &
$54.5$ &
$69.8$ &
$63.0$ &
$65.9$ &
$9.5$ &
$52.9$ & $50.5 ~ \scriptstyle{(0.0)}$ \\

~ \small \texttt{Foo Bar}  & $57.0$ &
$36.8$ &
$34.5$ &
$53.3$ &
$63.5$ &
$51.6$ &
$53.4$ &
$21.0$ &
$51.2$ & $46.9 ~ \scriptstyle{(-7.1)}$\\

~ \small ZS-CoT & $61.5$ &
$26.6$ &
$37.3$ &
$59.1$ &
$52.8$ &
$32.8$ &
$51.3$ &
$15.6$ &
$63.7$ & $44.5 ~\scriptstyle{(-11.9)}$\\

~ \small OPRO   & $77.1$ &
$43.1$ &
$44.4$ &
$81.0$ &
$75.5$ &
$64.8$ &
$\textbf{{73.1}}$ &
$36.5$ &
$62.7$ & $62.0 ~\scriptstyle{(22.8)}$ \\

~ \small OPRO-ICL & $81.6$ &
$\textbf{{44.1}}$ &
$43.8$ &
$68.8$ &
$74.5$ &
$65.9$ &
$73.0$ &
$36.6$ &
$70.8$ & $62.1 ~ \scriptstyle{(23.0)}$ \\

~ \small Random Vocabulary  &  $80.6$ &
$43.4$ &
$40.2$ &
$77.0$ &
$76.8$ &
$62.7$ &
$73.0$ &
$34.5$ &
$\textbf{\emphasize{72.2}}$ & $62.3 ~ \scriptstyle{(23.4)}$ \\

~ \small Random w/o Context &  $77.2$ &
$41.9$ &
$\textbf{\emphasize{45.7}}$ &
$75.5$ &
$\textbf{\emphasize{78.6}}$ &
$67.2$ &
$72.4$ &
$\textbf{\emphasize{39.5}}$ &
$66.8$ & $62.8 ~ \scriptstyle{(24.4)}$ \\

~  \small Random with Context &  $\textbf{\emphasize{82.0}}$ &
$43.9$ &
$42.9$ &
$\textbf{\emphasize{81.6}}$ &
$73.9$ &
$\textbf{\emphasize{69.5}}$ &
$71.4$ &
$35.8$ &
$71.2$ & $\textbf{\emphasize{63.6}} ~ \scriptstyle{(25.9)}$ \\

\midrule

\small \textsc{GPT2-XL 1.5B}   & & & & & & & & & & \\

~ \small \texttt{Answer:}  &  $72.3$ &
$37.7$ &
$38.3$ &
$69.5$ &
$60.8$ &
$59.2$ &
$61.2$ &
$7.2$ &
$46.5$ & $50.3 ~ \scriptstyle{(0.0)}$\\

~ \small \texttt{Foo Bar}  &  $40.3$ &
$40.5$ &
$40.4$ &
$49.5$ &
$56.4$ &
$47.7$ &
$56.6$ &
$17.1$ &
$57.0$ & $45.1 ~ \scriptstyle{(-10.3)}$ \\

~ \small ZS-CoT & $39.8$ &
$27.7$ &
$41.5$ &
$42.6$ &
$43.7$ &
$49.4$ &
$55.2$ &
$16.6$ &
$56.3$ & $41.4 ~\scriptstyle{(-17.7)}$\\

~ \small OPRO   &  $80.0$ &
$44.6$ &
$47.0$ &
$79.3$ &
$78.0$ &
$68.6$ &
$75.6$ &
$29.8$ &
$72.0$ & $63.9 ~ \scriptstyle{(27.0)}$ \\

~ \small OPRO-ICL  & $\textbf{{82.9}}$ &
$45.2$ &
$47.4$ &
$81.0$ &
$78.0$ &
$69.9$ &
$\textbf{{78.8}}$ &
$26.1$ &
$69.6$ & $\textbf{64.3} ~ \scriptstyle{(27.8)}$ \\

~ \small Random Vocabulary  &  $73.1$ &
$\textbf{\emphasize{45.9}}$ &
$\textbf{\emphasize{47.6}}$ &
$71.4$ &
$78.4$ &
$65.5$ &
$77.1$ &
$25.5$ &
$69.3$ & $61.5 ~ \scriptstyle{(22.3)}$ \\

~ \small Random w/o Context &  $72.0$ &
$40.5$ &
$44.6$ &
$75.0$ &
$\textbf{\emphasize{79.3}}$ &
$\textbf{\emphasize{70.8}}$ &
$72.8$ &
$\textbf{\emphasize{35.6}}$ &
$68.1$ & $62.1 ~ \scriptstyle{(23.5})$ \\

~  \small Random with Context &  $82.1$ &
$41.7$ &
$44.2$ &
$\textbf{\emphasize{81.8}}$ &
$78.5$ &
$64.3$ &
$73.4$ &
$32.9$ &
$\textbf{\emphasize{74.5}}$  & $63.7 ~ \scriptstyle{(26.6)}$ \\

\midrule

\small \textsc{Mistral 7B}   & & & & & & & & & & \\

~ \small \texttt{Answer:}  & $85.5$
& $46.6$
& $63.7$
& $89.0$
& $\textbf{92.2}$
& $61.6$
& $69.7$
& $34.3$
& $82.1$
 & $69.4 ~ \scriptstyle{(0.0)}$ \\

~ \small \texttt{Foo Bar}  & $61.0$
& $44.0$
& $63.7$
& $72.7$
& $87.1$
& $48.2$
& $55.9$
& $34.8$
& $79.5$
 &  $60.8 ~ \scriptstyle{(-12.4)}$\\
~ \small ZS-CoT  & 52.6 
& 44.5 
& 66.7
& 56.9
& 80.3
& 48.5
& 54.5
& 36.3
& 76.3
 &  $57.4 ~ \scriptstyle{(-17.3)}$\\
 ~ \small OPRO  & 86.3 
& 46.8
& 71.9
& 92.7
& 87.4
& $\textbf{79.5}$
& 82.7
& $\textbf{60.8}$
& $\textbf{83.1}$
 &  $76.8~ \scriptstyle{(10.7)}$\\
 ~ \small OPRO-ICL  & 91.0 
& 48.2
& 72.7
& $\textbf{93.5}$
& 90.6
& 76.2
& $\textbf{86.3}$
& 59.9
& 82.9
 &  $\textbf{77.9}~ \scriptstyle{(12.2)}$\\
~ \small Random Vocabulary &  87.3
& $\textbf{\emphasize{49.4}}$
& 70.1
& 93.1
& 85.8
& 76.2
& 80.7
& 55.7
& 81.6
 &  $75.5~ \scriptstyle{(8.8)}$\\
~ \small Random w/o Context &  91.4 
& 48.7
& $\textbf{\emphasize{73.8}}$
& 91.6
& 86.3
& 74.4
& 79.8
& 60.5
& 80.9
 &  $76.4~ \scriptstyle{(10.1)}$\\
 ~ \small Random with Context &  $\textbf{\emphasize{92.7}}$
& 47.9
& 74.1
& 92.9
& 87.8
& 67.0
& 84.4
& 59.9
& 82.9
 &  $76.6~ \scriptstyle{(10.4)}$\\

\midrule

\small \textsc{Mistral 7B Instruct}   & & & & & & & & & & \\

~ \small \texttt{Answer:}  & 86.5  
& 38.8
& 83.9
& 86.0
& 86.4
& 75.1
& 66.6
& 63.0
& 79.8
 &  $74.0 ~ \scriptstyle{(0.0)}$\\

~ \small \texttt{Foo Bar}  & 85.1  
& 39.7
& 82.3
& 85.3
& 85.7
& 75.5
& 63.8
& 63.8
& 78.8
 &  $73.3 ~ \scriptstyle{(-0.1)}$\\

~ \small ZS-CoT  &  83.8
& 39.8 
& 79.4
& 83.9
& 87.3
& 75.5
& 66.6
& 67.2
& 79.3
 &  $73.6 ~ \scriptstyle{(0.0)}$\\

~ \small OPRO  & 89.0
& 40.2
& 82.8
& 88.0
& 84.3
& $\textbf{81.3}$
& 68.3
& $\textbf{67.3}$
& 81.6
&  $75.9 ~ \scriptstyle{(2.6)}$\\

~ \small OPRO-ICL  &  89.1
& $\textbf{41.7}$
& 83.7
& $\textbf{90.2}$
& 87.7
& 79.6
& $\textbf{72.7}$
& 66.3
& $\textbf{82.3}$
 &  $\textbf{77.0} ~ \scriptstyle{(4.1)}$\\

~ \small Random Vocabulary & 87.0 
& 40.6
& $\textbf{\emphasize{84.0}}$ 
& 87.2 
& 87.6
& 78.8
& 66.6
& 66.5
& 79.1
&  $75.3 ~ \scriptstyle{(1.8)}$\\

 ~ \small Random w/o Context & 89.5 
& 40.9
& 82.4
& 88.2 
& 88.5
& 80.7
& 65.8
& 65.0
& 80.3
 &  $75.7 ~ \scriptstyle{(2.3)}$\\
 
 ~ \small Random with Context & $\textbf{\emphasize{89.6}}$ 
& 41.6
& 83.1
& 88.8
& $\textbf{\emphasize{89.6}}$ 
& 81.2
& 71.6
& 66.5
& 80.7
 & $\textbf{\emphasize{77.0}} ~ \scriptstyle{(4.1)}$ \\
 
\midrule

\small \textsc{LLaMA2 7B}   & & & & & & & & & & \\

~ \small \texttt{Answer:} & $83.0$ &
$43.0$ &
$68.1$ &
$89.0$ &
$89.1$ &
$67.6$ &
$63.6$ &
$35.5$ &
$80.2$ & $68.8 ~ \scriptstyle{(0.0)}$\\

~ \small \texttt{Foo Bar}  & $76.2$ &
$46.1$ &
$65.3$ &
$75.7$ &
$76.0$ &
$51.0$ &
$51.4$ &
$26.6$ &
$77.8$ & $60.7~ \scriptstyle{ (-11.8)}$ \\

~ \small ZS-CoT & $64.5$ &
$45.9$ &
$64.8$ &
$73.8$ &
$88.8$ &
$66.3$ &
$51.6$ &
$47.2$ &
$81.6$ & $64.9 ~ \scriptstyle{(-5.7)}$\\

~ \small OPRO &  $89.1$ &
$46.0$ &
$72.0$ &
$\textbf{{93.1}}$ &
$83.9$ &
$78.5$ &
$\textbf{{81.1}}$ &
$55.8$ &
$81.0$ & $75.6 ~ \scriptstyle{(9.9)}$ \\

~ \small OPRO-ICL & $92.0$ &
$48.8$ &
$\textbf{{72.3}}$ &
$92.5$ &
$85.6$ &
$\textbf{{79.9}}$ &
$79.1$ &
$\textbf{{57.3}}$ &
$80.7$ & $\textbf{76.5} ~ \scriptstyle{(11.2)}$ \\

~ \small Random Vocabulary  & $91.5$ &
$47.5$ &
$68.8$ &
$93.0$ &
$\textbf{\emphasize{88.4}}$ &
$79.2$ &
$77.1$ &
$49.4$ &
$80.7$ & $75.1 ~ \scriptstyle{(9.2)}$ \\

~ \small Random w/o Context & $\textbf{\emphasize{92.2}}$ &
$48.5$ &
$71.9$ &
$92.6$ &
$88.0$ &
$77.7$ &
$75.6$ &
$47.7$ &
$81.0$ & $75.0 ~ \scriptstyle{(9.0)}$ \\

~  \small Random with Context & $90.5$ &
$\textbf{\emphasize{49.5}}$ &
$71.6$ &
$\textbf{\emphasize{93.1}}$ &
$82.9$ &
$77.6$ &
$77.7$ &
$52.8$ &
$\textbf{\emphasize{82.2}}$ & $75.3 ~ \scriptstyle{(9.4)}$ \\

\midrule

\small \textsc{LLaMA2-Chat 7B}  & & & & & & & & & & \\

~ \small \texttt{Answer:}   & $85.3$ &
$38.3$ &
$34.8$ &
$87.4$ &
$82.3$ &
$79.5$ &
$58.8$ &
$48.0$ &
$70.5$ & $65.0 ~ \scriptstyle{(0.0)}$ \\

~ \small \texttt{Foo Bar}  & $85.9$ &
$29.8$ &
$36.6$ &
$80.3$ &
$83.9$ &
$78.3$ &
$53.3$ &
$38.9$ &
$61.8$ & $61.0 ~ \scriptstyle{(-6.2)}$ \\

~ \small ZS-CoT  & $82.3$ &
$26.1$ &
$34.3$ &
$77.2$ &
$79.8$ &
$75.9$ &
$52.0$ &
$34.9$ &
$58.4$ & $57.9 ~ \scriptstyle{(-10.9)}$\\

~ \small OPRO & $84.2$ &
$43.0$ &
$36.5$ &
$83.8$ &
$80.6$ &
$82.6$ &
$62.3$ &
$45.9$ &
$65.8$ & $65.0 ~ \scriptstyle{(0.0)}$ \\

~ \small OPRO-ICL & $85.5$ &
$45.2$ &
$39.2$ &
$89.4$ &
$83.8$ &
$83.5$ &
$\textbf{{65.6}}$ &
$49.3$ &
$71.2$ & $\textbf{68.1} ~ \scriptstyle{(4.8)}$ \\

~ \small Random Vocabulary  & $88.1$ &
$38.8$ &
$\textbf{\emphasize{39.5}}$ &
$88.8$ &
$\textbf{\emphasize{84.9}}$ &
$\textbf{\emphasize{83.6}}$ &
$58.0$ &
$\textbf{\emphasize{50.3}}$ &
$68.5$ & $66.7 ~ \scriptstyle{(2.6)}$ \\

~ \small Random w/o Context & $\textbf{\emphasize{90.2}}$ &
$42.3$ &
$38.4$ &
$89.3$ &
$83.7$ &
$82.2$ &
$60.3$ &
$48.8$ &
$\textbf{\emphasize{72.3}}$ & $67.5 ~ \scriptstyle{(3.8)}$ \\

~  \small Random with Context & $89.6$ &
$\textbf{\emphasize{47.5}}$ &
$37.6$ &
$\textbf{\emphasize{89.7}}$ &
$83.7$ &
$82.0$ &
$63.8$ &
$48.8$ &
$66.9$ & $67.7 ~ \scriptstyle{(4.2)}$ \\

\midrule
\small \textsc{ChatGPT (GPT-3.5)}  &    &    &     &     &     &      &      &    &    & \\

~ \small \texttt{Answer:} & $93.2$ &
$44.1$ &
$90.2$ &
$91.8$ &
$90.2$ &
$68.0$ &
$78.9$ &
$75.0$ &
$81.6$ & $79.2 ~ \scriptstyle{(0.0)}$ \\

~ \small \texttt{Foo Bar}  & $91.0$ &
$37.1$ &
$90.0$ &
$88.1$ &
$89.1$ &
$71.5$ &
$66.0$ &
$72.3$ &
$81.2$ & $76.3 ~ \scriptstyle{(-3.7)}$\\

~ \small ZS-CoT & $93.9$ &
$35.0$ &
$87.9$ &
$91.2$ &
$89.8$ &
$76.0$ &
$80.1$ &
$76.4$ &
$82.6$ & $79.2 ~ \scriptstyle{(0.0)}$ \\

~ \small OPRO  & $93.6$ &
$43.8$ &
$89.5$ &
$91.4$ &
$87.9$ &
$80.7$ &
$80.5$ &
$72.1$ &
$83.4$ & $80.3 ~ \scriptstyle{(1.4)}$\\

~ \small OPRO-ICL  & $94.3$ &
$36.3$ &
$90.8$ &
$90.8$ &
$90.2$ &
$82.4$ &
$78.9$ &
$75.0$ &
$83.0$ & $80.2 ~ \scriptstyle{(1.3)}$\\

~ \small Random Vocabulary   & $\textbf{\emphasize{94.7}}$ &
$47.7$ &
$89.5$ &
$91.8$ &
$\textbf{\emphasize{91.6}}$ &
$81.8$ &
$78.9$ &
$\textbf{\emphasize{77.7}}$ &
$83.0$ & $81.9 ~ \scriptstyle{(3.4)}$\\

~ \small Random w/o Context & $93.4$ &
$42.4$ &
$\textbf{\emphasize{91.6}}$ &
$91.2$ &
$90.8$ &
$82.6$ &
$79.5$ &
$74.6$ &
$82.4$ & $80.9 ~ \scriptstyle{(2.1)}$\\

~  \small Random with Context  & $94.3$ &
$\textbf{\emphasize{48.4}}$ &
$89.3$ &
$\textbf{\emphasize{92.6}}$ &
$87.3$ &
$\textbf{\emphasize{84.4}}$ &
$\textbf{\emphasize{84.0}}$ &
$75.2$ &
$\textbf{\emphasize{83.6}}$ & $\textbf{\emphasize{82.1}} ~ \scriptstyle{(3.7)}$\\

\bottomrule[1pt]
\end{tabular}
\endgroup

}

\end{center}

\caption{Our main results on the evaluation set. We use one-shot context for all experiments, and use the same model for separator generation and evaluation. The relative improvement scores are computed using the ``\texttt{Answer:}'' as baseline. All the results except ChatGPT are calculated based on five different random seeds. For ChatGPT, we use two different random seeds. Results are colored blue when our random separators achieve the best performance.}
\label{tab:main_result}
\end{table*}

\paragraph{Prompting settings.} We use one-shot examples as context to prompt language models during both training and test stages. When context is necessary for separator generation, we provide three randomly chosen training examples. We set the generation temperature to 1.0 and use a temperature of 0.0 for prompt-based text classification. 
For training of OPRO and OPRO-ICL, we set a maximum of 40 optimisation steps and generate four candidate separators each step. For our random methods, we generate 160 candidate separators and select the best one for evaluation.

\section{Results and Discussion}
We report our results in Table~\ref{tab:main_result} and demonstrate the effectiveness of randomly sampled separators across all tasks. In addition, we compare four additional baseline methods in Table~\ref{tab:more-models} using the experimental settings described by~\citet{guo2023connecting-evoprompt}.

\begin{table}[!ht]
\setlength{\tabcolsep}{2.5pt}
\begin{center}
\centering
\resizebox{\columnwidth}{!}{
\renewcommand{\arraystretch}{1.0}
\begin{tabular}{lcccccccc}
\toprule[1pt]
           & \begin{tabular}{c}
                \small MI
           \end{tabular}  & \begin{tabular}{c}
                \small NI\\
           \end{tabular} &  \begin{tabular}{c}
                \small APE \\
           \end{tabular} & \begin{tabular}{c}
                \small Evo \\
                \small Prompt-DE
           \end{tabular}  & \begin{tabular}{c}
                \small Evo \\
                \small Prompt-GA
           \end{tabular}  & \begin{tabular}{c}
                \small Random\\
                \small Vocabulary
           \end{tabular}  & \begin{tabular}{c}
                \small Random\\
                \small w/o Context
           \end{tabular} & \begin{tabular}{c}
                \small Random \\
                \small w/Context
           \end{tabular} 
           \\
\midrule[1pt]
\small SST-2  & 93.7 & 92.9 & 94.0 & 94.8 & 94.8 & 93.7 & 94.2& 94.4\\
\small CR  & 91.4 &  90.9 & 90.5 & 91.4 &91.2& 91.1 &  90.2 &   91.0\\
\small MR & 88.8 & 89.6 & 90.9 & 90.2 &  90.4 & 89.4 & 89.6 & 90.3\\
\small SST-5  & 42.9 & 48.6 & 47.0 & 48.2 & 49.4  &  41.0& 37.1 & 45.5\\
\small AGNews & 70.6  & 48.9 & 71.2 & 73.3 & 73.4 & 76.5 & 80.6 & 79.5\\
\small TREC & 50.6 & 55.0 & 59.6 & 64.4 & 63.8 & 61.4 & 57.6 & 66.8\\
\small SubJ & 49.8 & 52.6 & 63.3 & 77.6 & 67.9 & 62.6 & 69.0 & 64.7\\
\midrule[1pt]
\small \textsc{AVG.} & 71.1 & 68.2 & 73.8 & 77.1 & 75.9 & 73.7 & 74.0 & 76.0\\

\bottomrule[1pt]
\end{tabular}
}
\end{center}
\caption{Prompt performance on the Alpaca-tuned LLaMA model. We compare with two human-level prompt optimisation methods, MI~\cite{zhang2023sentiment-mi} and NI~\cite{mishra2021cross-ni}, and two automatic prompt optimisation methods, APE~\cite{zhou2022large-ape} and EvoPrompt~\cite{guo2023connecting-evoprompt}.} 
\label{tab:more-models}
\end{table}

\subsection{Random Separators are Strong Baselines}

\paragraph{Unnatural separators are effective.} To our surprise, we find that even separators chosen at random from the vocabulary can substantially improve performance. Table~\ref{tab:main_result} shows that the \textit{Random Vocabulary} method yields, on average, a 10\% relative improvement across nine benchmark datasets compared to human-curated separators. 
\textit{Random Vocabulary} shows only marginal differences~(less than 1\% difference) compared to self-optimisation style methods~\citep{yang2023large-opro,zhou2022large-ape}.
For evolutionary methods~\cite{guo2023connecting-evoprompt}, \textit{Random Vocabulary} shows a 3.4\% difference from the best EvoPrompt result~(Table~\ref{tab:more-models}), with the most significant drop~(2.1\%) occurring on the SubJ dataset.

Since our goal is not to produce a state-of-the-art method, but rather to establish a strong baseline, the performance of \textit{Random Vocabulary} suggests that previous progress and the weight of contributions in prompt optimisation might be overestimated.
Notably, our method does not depend on a language model for generation~(Table~\ref{tab:three-random-method-compare}).
This result challenges the common practice in prompt optimisation where large language models are used for creating alternative prompts.

\paragraph{Natural language separators are effective, but may not be essential.} Human creation of separators often takes coherence into account. For instance, in sentiment classification, a model might struggle to predict ``positive'' or ``negative'' due to high sequence perplexity, but introducing a separator like ``It is positive'' can align predictions with the model's pre-training objective. Given the surprisingly competitive quality of \textit{Random Vocabulary}, we further explore the potential benefits of coherence towards prompt optimisation. 
We design another simple strategy, \textit{Random w/o Context}, which samples from the language model's prior to generate natural language phrases as separators. The key difference between \textit{Random w/o Context} and \textit{Random Vocabulary} is that the former consists of natural language phrases, whereas the latter may not. Both methods generate separators that are statistically unlikely to be relevant to the task and context. 

Experimental results show that the \textit{Random w/o Context} is significantly better than human baselines~(12\% relative improvement) and nearly on par with previous state-of-the-art prompt optimisation methods~(less than a 1\% difference). This suggests that \textit{Random w/o Context} is also a simple but strong baseline for prompt optimisation. 
When comparing the natural and unnatural separators, our analysis shows a mere 0.5\% relative difference between \textit{Random Vocabulary} and \textit{Random w/o Context}. Such a small margin suggests that, while the \textit{Random w/o Context} approach is competitive, coherence does not appear to be a critical factor.

\paragraph{Task information in separator generation provides slight improvements.} \textit{Random with Context} imposes task-relevance constraints by sampling from a language model conditioned on training samples. We observe, though marginal, consistent improvements of including context information for prompt optimisation. Specifically, the \textit{Random with Context} method achieves a relative improvement of 0.3\% over \textit{Random w/o Context} and 0.9\% over \textit{Random Vocabulary}.
Nevertheless, given the significant gains that \textit{Random Vocabulary} achieves over human baseline~(10\%), the incremental gains from adding task information are relatively minor.

\subsection{{Random Sampling is a Strong Prompt Optimiser}}
\paragraph{Random separator generation methods are comparable to instruction-based approaches.} 
OPRO and its in-context learning variant achieve the top average performance for four out of seven models. However, the advantage is minimal, with a 0.1\% average performance difference compared to random methods. It appears that instruction-based methods might be randomly encountering good separators throughout the optimisation process.

\paragraph{Instruction-tuned models are not essential for proposing separators.} Previous prompt optimisation work heavily rely on a language model to suggest alternatives, and they make a strong assumption that only instruction-tuned models are able to differentiate good and bad prompts. However, our observation indicates that instruction-tuned models are not essential for proposing the separators. As shown in Table~\ref{tab:main_result}, OPRO is able to achieve reasonable performance, even for the smaller GPT2 family models.  It is worth noting that, given the linguistic complexity of the OPRO-style meta prompt, a GPT2 model is unlikely to ``understand'' it~\cite{ouyang2022training-instructgpt}.
Furthermore, we do not observe a significant increase when using instruction-based model for proposing separators. 

\begin{table}[!htb]
\setlength{\tabcolsep}{2pt}
\begin{center}
\centering
\resizebox{1.0\columnwidth}{!}{
\renewcommand{\arraystretch}{0.95}

\begin{tabular}{llc}
\toprule
Strategy & Separator & Score \\
\midrule

\small OPRO & \small \texttt{The new text is the following:} & 92.2\\
\small OPRO-ICL & \small \texttt{00:57} & 92.2\\
\small Random Vocabulary &  
\small \texttt{obliged\symbol{92}u0442\symbol{92}u0438\symbol{92}u0435Circ song} & 92.2\\
\small Random w/o Context &  \small \texttt{Home Business New \symbol{92}u2018} & 92.2\\
\small  Random with Context &  \small \texttt{**GW - The Wall Street} & 92.2\\

\bottomrule
\end{tabular}
}
\end{center}
\caption{Performant separators discovered in the training process on AGNews using \textsc{LLaMA2 7B}, we report the accuracy score over the training set.} 
\label{tab:case-study}
\end{table}

\begin{table}[!ht]
\begin{center}
\centering
\resizebox{1.0\columnwidth}{!}{
\renewcommand{\arraystretch}{0.95}

\begin{tabular}{lccc}
\toprule
           & \begin{tabular}{c}
                \small Random\\
                \small Vocabulary
           \end{tabular}  & \begin{tabular}{c}
                \small Random\\
                \small w/o Context
           \end{tabular}  & \begin{tabular}{c}
                \small Random\\
                \small with Context
           \end{tabular}  \\
\midrule
\small \textsc{GPT2-Large} & 37.1\% & 20.4\% & 51.2\%  \\
\small \textsc{GPT2-XL}  &  70.6\% & 42.6\% & 48.3\% \\
\small \textsc{LLAMA2 7B}  & 66.1\% & 47.2\% & 49.6\% \\
\small \textsc{LLAMA2 Chat}  & 21.4\% & 14.6\% & 13.9\% \\
\bottomrule
\end{tabular}
}
\end{center}
\caption{Chance of random separators outperforming the human baseline ``\texttt{Answer:}'' on the AGNews dataset.}
\label{tab:good-separator-ratio}
\end{table}

\section{Analyses}
\subsection{Language Space is Rich with Potentially Good Separators.}\label{sec:language-rich-good}

We find that different approaches can discover distinct separators while yielding similar performance, as shown in Table~\ref{tab:case-study}. This prompts a natural question: how many effective separators exist? For simplicity, we deem any separator effective if it outperforms a human-curated separator such as ``Answer:''. To investigate this question, we calculate the percentage of effective random separators based on all data points\footnote{Approximately 10,000 separators.} from the main experiment.
Table~\ref{tab:good-separator-ratio} shows that our random baseline has on average a 40\% chance to draw a separator that is better than the human baseline. 
This suggests that in the language space, there are more performant separators than we previously expected.

\subsection{Are Performant Random Separators Transferable?}

In this section, we study whether performant separators discovered by random approaches are transferable across different tasks and contexts. 

\paragraph{Cross-task transferability.} We use separators with the highest accuracy from each task~(Table~\ref{tab:transfer-text}) and apply them to different tasks. According to Table~\ref{tab:transfer_task}, we do not observe high transferability across tasks; for example, a performant prompt for SST2 has almost random-guessing performance on SST5, and vice versa. This is within our expectations, as these random prompts are optimised and selected for a particular task. Notably, even widely used human-curated separators exhibit similar transferability scores~(59.0\% versus 59.4\% in average score) across tasks to random methods.

\definecolor{severe}{RGB}{241,113,31}
\definecolor{severe-moderate}{RGB}{215,75,63}
\definecolor{moderate}{RGB}{177,50,90}
\definecolor{moderate-light}{RGB}{135,33,107}
\definecolor{light}{RGB}{180,180,180}

\begin{table}[!t]
\setlength{\tabcolsep}{2.5pt}
\begin{center}
\centering
\resizebox{0.9\columnwidth}{!}{%
\begingroup
\renewcommand{\arraystretch}{0.9} 
\begin{tabular}{lccccccccccc}
\toprule[1pt]
      \multirow{2}{*}{\qquad$\swarrow$} &
  \multirow{2}{*}{SST-2} &
  \multirow{2}{*}{SST-5} &
  \multirow{2}{*}{DBPedia} &
  \multirow{2}{*}{MR} &
  \multirow{2}{*}{CR} &
  \multirow{2}{*}{MPQA} &
  \multirow{2}{*}{Subj} &
  \multirow{2}{*}{TREC} &
  \multirow{2}{*}{AGNews} &
  \multirow{2}{*}{\begin{tabular}[c]{@{}c@{}}Avg. \end{tabular}} \\
 &
   &
   &
   &
   &
   &
   &
   &
   &
   &
   \\
\midrule[1pt]
\multicolumn{11}{l}{\small \textsc{Best Human-level Prompt}} \\
\small \texttt{Answer:} & 81.2 & 40.6 & 40.6 & {82.8} & 81.2& 76.6 & \textbf{79.7} & 4.7 & 43.8 & 59.0 \\
\midrule
\multicolumn{11}{l}{\small \textsc{Best Separator, Random Vocabulary}} \\

SST2 & \textcolor{light}{{85.9}} & \textcolor{moderate}{39.1} & 42.2 & 43.8 & \textcolor{moderate}{81.2} & \textcolor{severe}{84.4} & \textcolor{moderate}{65.6} & 26.6 & \textcolor{moderate}{65.6} & 59.4 \\
SST5 & 57.8 & \textcolor{light}{45.3} & 31.2 & \textcolor{moderate}{73.4} & \textcolor{moderate}{81.2} & 48.4 & 51.6 & 7.8 & 56.2 & 50.3 \\
DBPedia & 50.0 & \textcolor{moderate}{39.1} & \textcolor{light}{56.2} & 64.1 & \textcolor{moderate}{81.2} & \textcolor{severe}{79.7} & 42.2 & 29.7 & 56.2 & 55.4 \\
MR & \textcolor{moderate}{75.0} & 34.4 & 39.1 & \textcolor{light}{{82.8}} & \textcolor{moderate}{82.8} & 60.9 & 42.2 & 21.9 & 48.4 & 54.2 \\
CR & 48.4 & 29.7 & 37.5 & 43.8 & \textcolor{light}{{93.8}} & 62.5 & 42.2 & 28.1 & 57.8 & 49.3 \\
MPQA & \textcolor{severe}{85.9} & \textcolor{moderate}{39.1} & 42.2 & 43.8 & \textcolor{moderate}{81.2} & \textcolor{light}{{84.4}} & \textcolor{moderate}{65.6} & 26.6 & \textcolor{moderate}{65.6} & 59.4 \\
SubJ & 57.8 & 25.0 & \textcolor{moderate}{46.9} & \textcolor{severe}{76.6} & \textcolor{severe}{87.5} & \textcolor{moderate}{73.4} & \textcolor{light}{76.6} & 28.1 & 48.4 & 57.8 \\
TREC & 57.8 & 23.4 & 39.1 & \textcolor{severe}{76.6} & \textcolor{moderate}{81.2} & \textcolor{severe}{76.6} & 42.2 & \textcolor{light}{43.8} & 60.9 & 55.7 \\
AGNews & 50.0 & 32.8 & 31.2 & \textcolor{severe}{79.7} & \textcolor{moderate}{79.7} & \textcolor{severe}{76.6} & 42.2 & 25.0 & \textcolor{light}{79.7} & 55.2 \\

\midrule

\multicolumn{11}{l}{\small \textsc{ Best Separator, Random w/o Context}} \\
SST2 & \textcolor{light}{76.6} & 18.8 & 40.6 & 57.8 & \textcolor{severe}{93.8} & 59.4 & 56.2 & 34.4 & 59.4 & 55.2 \\
SST5 & 51.6 & \textcolor{light}{43.8} & 31.2 & 62.5 & \textcolor{severe}{85.9} & \textcolor{severe}{75.0} & \textcolor{moderate}{67.2} & \textcolor{moderate}{37.5} & 29.7 & 53.8 \\
DBPedia & 51.6 & 29.7 & \textcolor{light}{54.7} & 54.7 & 70.3 & 60.9 & 60.9 & \textcolor{severe}{43.8} & \textcolor{severe}{71.9} & 55.4 \\
MR & 60.9 & \textcolor{moderate}{35.9} & \textcolor{moderate}{43.8} & \textcolor{light}{{84.4}} & \textcolor{moderate}{81.2} & \textcolor{severe}{76.6} & 48.4 & 31.2 & 45.3 & 56.4 \\
CR & \textcolor{severe}{76.6} & 18.8 & 40.6 & 57.8 & \textcolor{light}{{93.8}} & 59.4 & 56.2 & 34.4 & 59.4 & 55.2 \\
MPQA & 56.2 & 23.4 & \textcolor{severe}{50.0} & 43.8 & \textcolor{moderate}{79.7} & \textcolor{light}{{81.2}} & 42.2 & \textcolor{moderate}{37.5} & 46.9 & 51.2 \\
SubJ & 50.0 & 18.8 & \textcolor{moderate}{48.4} & 59.4 & \textcolor{severe}{89.1} & \textcolor{moderate}{68.8} & \textcolor{light}{{81.2}} & 17.2 & 23.4 & 50.7 \\
TREC & 50.0 & 20.3 & \textcolor{moderate}{48.4} & 57.8 & \textcolor{moderate}{79.7} & \textcolor{severe}{73.4} & 42.2 & \textcolor{light}{45.3} & 54.7 & 52.4 \\
AGNews & 48.4 & 25.0 & 42.2 & 60.9 & \textcolor{moderate}{79.7} & \textcolor{moderate}{68.8} & 42.2 & 28.1 & \textcolor{light}{79.7} & 52.8 \\

\bottomrule[1pt]
\end{tabular}
\endgroup

}

\end{center}

\caption{Random separator transferability test on GPT2-XL. We transfer the best random separator from each task~(the columns) to the others~(the rows), then colour the results according to their \textit{relative} accuracy on the training set. Brightness denotes high transferability, with thresholds at 80\% and 90\%.}
\label{tab:transfer_task}
\end{table}

\begin{table}[!htb]
\begin{center}
\centering
\resizebox{0.96\columnwidth}{!}{
\renewcommand{\arraystretch}{0.96}
\begin{tabular}{lcc}
\toprule
Task & Random Vocabulary & Random w/o Context \\
\midrule

SST2 &  ``\texttt{cancell BlakesteamappsGr}'' & ``\texttt{In December, I}''\\
SST5 & ``\texttt{biblical namely}'' &  ``\texttt{(}''\\
DBPedia &  ``\texttt{download pitch Par}''
 & ``\texttt{To view this}''\\
MR & ``\texttt{GAME paced}'' & ``\texttt{"}''\\
CR & ``\texttt{learnt}'' & ``\texttt{In December, I}''\\
MPQA & ``\texttt{cancell BlakesteamappsGr}'' & ``\texttt{LONDON}''\\
SubJ & ``\texttt{pubfile Favor}'' & ``\texttt{A small}''\\
TREC & ``\texttt{wasSIZE Armageddon}'' & ``\texttt{Image}''\\
AGNews & ``\texttt{Alc messenger SYSTEM precipitation}'' & ``\texttt{Weird Al}''\\

\bottomrule
\end{tabular}
}
\end{center}
\caption{Best-discovered random separators used in the transferability test.} 
\label{tab:transfer-text}
\end{table}

\begin{table}[!t]
\begin{center}
\centering
\resizebox{1.0\columnwidth}{!}{
\begingroup
\renewcommand{\arraystretch}{1.0}
\begin{tabular}{lcccccc}
\toprule
  \multirow{1}{*}{~} &
  \multirow{1}{*}{\#1} &
  \multirow{1}{*}{\#2} &
  \multirow{1}{*}{\#3} &
  \multirow{1}{*}{\#4} &
  \multirow{1}{*}{\#5} &
  \multirow{1}{*}{Avg.}\\
\midrule
\multicolumn{6}{l}{\small \textsc{Best Human-level Prompt}} \\
\small \texttt{Answer:} & 43.8 &
54.7 &
57.8 &
43.8 &
53.1 & 
50.6 \\
\midrule
\multicolumn{6}{l}{\small \textsc{Average Separators, Random Vocabulary}} \\
\small \textsc{Avg.} \texttt{Foo Bar} & 55.7 &
55.6 &
63.2 &
61.9 &
53.3 &
57.9 \\
\midrule
\multicolumn{6}{l}{\small \textsc{Best Separator, Random Vocabulary}} \\

\small\texttt{\#1 Best Sep.} & \textcolor{light}{79.7} & \textcolor{severe}{81.3} & \textcolor{severe}{81.3} & \textcolor{moderate}{68.8} & 53.2 & 72.9 \\
\small\texttt{\#2 Best Sep.} & \textcolor{severe}{71.9} & \textcolor{light}{78.1} & \textcolor{severe}{79.7} & 64.1 & \textcolor{severe}{73.4} & 73.4 \\
\small\texttt{\#3 Best Sep.} & \textcolor{severe}{71.9} & 60.9 & \textcolor{light}{{82.8}} & \textcolor{moderate}{70.3} & \textcolor{severe}{76.6} & 72.5 \\
\small\texttt{\#4 Best Sep.} & \textcolor{severe}{76.6} & \textcolor{severe}{73.4} & \textcolor{moderate}{67.2} & \textcolor{light}{{82.8}} & 57.8 & 71.6 \\
\small\texttt{\#5 Best Sep.} & \textcolor{severe}{71.9} & \textcolor{moderate}{67.2} & \textcolor{moderate}{71.9} & \textcolor{moderate}{70.3} & \textcolor{light}{79.7} & 72.2 \\

\bottomrule
\end{tabular}
\endgroup

}

\end{center}

\caption{Random separator context transferability test on GPT2-XL. We choose the best separator from different contexts of the AGNews dataset; then compute its accuracy across other contexts for the same AGNews training set. \textsc{AVG.} \texttt{Foo Bar} represents the average performance of 160 randomly sampled separators. 
}
\label{tab:transfer_context}
\end{table}

\paragraph{Cross-context transferability.} Given the limited cross-task transferability of separators, we further study whether performant random separators are transferable when the context changes\footnote{The term ``context'' here is also referred to as ``demonstrations'' in in-context learning. In our experiments, we use one-shot examples as context to guide the output label space for classification tasks.} within a fixed task. We discover that random separators exhibit a degree of transferability and are significantly better than human-curated ones~(73.4\% versus 50.6\%). Surprisingly, for this task, human-curated prompts perform even worse than the average random prompt, which matches with our observations in Section~\ref{sec:language-rich-good}. Overall, our random strategies offer considerable flexibility in discovering task-wide performant separators.

\subsection{Beyond Text Classification Tasks}\label{sec:generative-tasks}
In previous sections, we have mainly showed that random sampling is a strong baseline for classification tasks across nine classification datasets over eight different models, revealing that the over-sensitivity of LLMs is still a notable issue, and it is a fundamental characteristic of in-context learning. 

To verify whether such random methods have similar patterns, and are strong baselines for generative reasoning tasks, we apply our ``random sampling over vocabulary'' method to GSM8K~\citep{cobbe2021training-gsm8k}, a mathematical reasoning dataset. 
\begin{table}[!ht]
\begin{center}
\centering
\resizebox{1.0\columnwidth}{!}{
\renewcommand{\arraystretch}{1.0}

\begin{tabular}{lccc}
\toprule
  \textsc{seed}  & \begin{tabular}{c}
                \small \textsc{Human CoT}
           \end{tabular}  & \begin{tabular}{c}
                \small \textsc{AVG.} \\
                \small \textsc{Random Vocabulary}\\
                \small 
           \end{tabular}  & \begin{tabular}{c}
                \small \textsc{Best} \\
                \small \textsc{Random Vocabulary} \\
                \small \textsc{(Rel. $\Delta\%$)}
           \end{tabular} \\
\midrule
\small \textsc{\#1} & 35.9 & 36.3 & 46.9~$\scriptstyle{(30.6)}$ \\
\small \textsc{\#2} & 42.2 & 39.3 & 50.0~$\scriptstyle{(18.5)}$ \\
\small \textsc{\#3} & 35.9 & 37.7 & 46.9~$\scriptstyle{(30.6)}$ \\
\small \textsc{\#4} & 39.1 & 38.0 & 45.3~$\scriptstyle{(15.9)}$\\
\hline
\small \textsc{AVG.} & 38.3 & 37.8 & 47.3 \\

\bottomrule
\end{tabular}
}
\end{center}
\caption{Comparing best random separators performance with Chain-of-Thoughts prompting on GSM8K.}
\label{tab:generative-task}
\end{table}

Similar to the previous experimental setup, we sample up to 160 different random separators, and then perform selection and evaluate their performance over the subset of test data.
For this task-specific setup, we use 5-shot and majority voting@1 on the Mistral 7B model~\cite{jiang2023mistral}, and report results on four different random seeds\footnote{We change the demonstration examples for each seed.}, as shown in Table~\ref{tab:generative-task}.

\paragraph{CoT is only slightly better than average random separators.} Surprisingly, for the few-shot mathematical reasoning task, thinking-style phrases could only be slightly better than random separators. On average, CoT attains an accuracy of 38.3 and the average of random separators is 37.8. This suggests that the gains of manual prompt optimisation are sub-optimal. 

\paragraph{CoT shows high variance in quality across different examples.}
As shown in Table~\ref{tab:generative-task}, we observe that the language model is sensitive to different sets of demonstrations across all methods.
The most widely used human-derived chain-of-thought prompt (``let's think step by step'') results in the highest variance,  with over a 17\% relative performance gap between the best and worst set of demonstrations. On the other hand, our random sampling approach yields a 9\% relative difference, suggesting better robustness.

\paragraph{Random separators are still a strong baseline for generative reasoning tasks.} Our best random separators reach an average accuracy of 47.3, a 23\% relative increase in accuracy over the CoT baseline. This aligns with our main findings derived from the classification tasks (Table~\ref{tab:main_result}). It is conceivable that for complex generative tasks, the language space is abundant with potentially good separators. 

\section{Related Work}

Automatically discovering effective prompts remains a challenging research problem because of the complexity of the search space. 
One direction is continuous prompt tuning~\cite{qin2021learning,lester2021power,liu2023gpt}, which involves adding a set of smaller tunable parameters to pretrained language models. An alternative approach is to optimise discrete token spaces. \citet{shin2020autoprompt} showed that gradient information at the embedding layer can guide the discovery of more effective prompts. According to their research, unnatural prompts can also result in good performance, which matches our observations. In spite of the efficiency of using gradient information, such a method, which heavily relies on the availability of language models, imposes some restrictions on certain types of models. 
An alternative direction is black box search. To simplify language space optimisation, \citet{prasad-etal-2023-grips} introduced a set of operations, such as add/delete/replace tokens. APE~\cite{zhou2022large-ape} showed that generating some alternatives and then selecting and rephrasing them could also provide effective solutions. Similarly, \citet{xu-etal-2022-gps} used evolutionary methods to optimise the search process. Recently, \citet{yang2023large-opro} demonstrated that we can teach language models to learn the pattern of good prompts using human-written meta prompts. EvoPrompt~\cite{guo2023connecting-evoprompt} showed how we can formulate the evolutionary process in meta-prompt. \citet{fernando2023promptbreeder} further demonstrated how we can improve the meta-prompt using language models, making the whole framework completely automatic without relying on the internal state of language models. 

\section{Conclusion}
We find that random separators, even those selected at random from vocabulary, could be as effective as previously discovered state-of-the-art prompts. In addition, we conduct research on three different types of random separators, which demonstrated that these random separators do not require instruction-tuned models, could provide a 12\% relative improvement as compared to human baselines, and are on par with a self-optimising approach involving complex meta-prompt engineering.

\section*{Limitations}
While we have done our utmost to explore randomly generated prompts, a limitation of this work is that we mainly evaluate our approaches on text classification tasks. 
However, based on our experimental results in the mathematical generative task~(Section~\ref{sec:generative-tasks}), our findings are still empirically sound. We will leave a more comprehensive evaluation of generative tasks as future work.

\section*{Acknowledgements}
We thank Jean Kaddour for his valuable feedback. Pontus Stenetorp would like to acknowledge the helpful proofing feedback from several viewers while finalising the submission. 
This work is supported by Microsoft Research via Accelerate Foundation Models Research Grant.

\bibliography{custom}
\bibliographystyle{acl_natbib}

\end{document}